\documentclass[11pt]{article}

\usepackage[]{acl}

\usepackage{times}
\usepackage{latexsym}

\usepackage[T1]{fontenc}

\usepackage[utf8]{inputenc}

\usepackage{microtype}

\usepackage{inconsolata}

\usepackage{graphicx}

\usepackage[utf8]{inputenc} 
\usepackage[T1]{fontenc}    
\usepackage{hyperref}       
\usepackage{url}            
\usepackage{booktabs}       
\usepackage{amsfonts}       
\usepackage{nicefrac}       
\usepackage{microtype}      
\usepackage{xcolor}         

\usepackage{times}
\usepackage{graphicx}
\usepackage{amsmath}
\usepackage{amssymb}

\usepackage{multirow}
\usepackage{array}
\usepackage{booktabs}
\usepackage{color}
\usepackage{xcolor}
\usepackage{pifont}
\usepackage{mathtools}
\usepackage{colortbl}
\usepackage{xcolor}
\usepackage{array}
\usepackage{color}
\usepackage{bbding}
\usepackage{makecell}
\usepackage{subfigure}
\usepackage{float}
\usepackage{wrapfig}
\usepackage{caption}
\usepackage{tcolorbox}
\usepackage{graphicx}
\usepackage{subcaption}
\usepackage[ruled,vlined]{algorithm2e}
\usepackage{tikz}
\usetikzlibrary{tikzmark}
\usepackage{bm}
\usepackage{listings}
\usepackage{placeins} 

\definecolor{mydarkgreen}{HTML}{009901}

\title{MGPO: Thinking with Images via Multi-Turn \\ Grounding-Based Reinforcement Learning}

\author{
\textbf{Xinyu Huang$^{1,2,*}$ \quad Yuhao Dong$^{2,*}$ \quad  Weiwei Tian$^{1}$ \quad Bo Li$^{2}$    \quad Rui Feng$^{1}$  \quad Ziwei Liu$^{2,\dagger}$}  \\
\normalsize
$^{1}$Fudan University \quad $^{2}$S-Lab, Nanyang Technological University \\
\small{$^{*}$Equal contribution. \quad $^{\dagger}$Corresponding author.}
}

\begin{document}
\maketitle
\begin{abstract}
    State-of-the-art large multi-modal models (LMMs) face challenges when processing high-resolution images, as these inputs are converted into enormous visual tokens, many of which are irrelevant to the downstream task. In this paper, we propose \textbf{Multi-turn Grounding-based Policy Optimization (MGPO)}, an end-to-end reinforcement learning~(RL) framework that enables LMMs to iteratively focus on key visual regions by automatically cropping sub-images, based on model-predicted grounding coordinates within a multi-turn conversation framework. Compared to supervised fine-tuning~(SFT), which requires costly additional grounding annotations, \textit{our approach highlights that LMMs can emerge robust grounding abilities during the RL training process, leveraging only a binary reward function derived from the correctness of the final answer}. Additionally, we observe that LMMs struggle to autonomously trigger visual grounding during the
rollout process. To address this cold start problem, we design a multi-turn conversational template and restrict policy loss computation to model outputs generated across multiple dialogue rounds, thereby promoting stable optimization. Extensive experiments demonstrate that, when trained on standard visual-question-short answering data without grounding annotations, MGPO effectively elicits stronger grounding capabilities compared to GRPO, leading to 5.4\% improvement on in-distribution MME-Realworld and 5.2\% improvement on the challenging out-of-distribution~(OOD) V* Bench. Notably, MGPO post-training on Qwen2.5-VL-7B with 21K samples surpasses OpenAI’s o1 and GPT-4o models on the OOD V* Bench. Codes are available at \textcolor{pink}{\url{https://github.com/EvolvingLMMs-Lab/MGPO}}.
\end{abstract}
\section{Introduction}


State-of-the-art large multimodal model (LMM) architectures, such as Qwen2.5-VL~\cite{bai2025qwen25_vl}, are typically built upon a powerful large language model (LLM) backbone (e.g., Qwen2.5~\cite{yang2024qwen2.5}) integrated with an external Native Resolution Vision Transformer (NaViT) module~\cite{dehghani2023navit,tschannen2025siglip2}. This module allows LMMs to process images at their original resolutions, enabling the model to capture fine-grained visual details and achieve high perceptual fidelity~\cite{liu2024llava_1_5,li2024llava-ov}.

Nevertheless, this approach also presents challenges in high-resolution real-world scenarios. {\it{(1)}} The number of visual tokens increases quadratically with image resolution, resulting in a large proportion of tokens that may be irrelevant to the downstream task, an issue analogous to the "needle-in-a-haystack" problem in LLMs~\cite{liu2023lost}. {\it{(2)}} Due to the inherent context length limitations of LMMs, a maximum pixel constraint is imposed on input images in practical applications, necessitating to resize images that exceed this threshold.

Inspired by the human visual system, although its theoretical resolution is estimated to be about 576 megapixels~\cite{ClarkVision2014,Curcio1990}, sharp vision is limited to the foveal area of the macula, which covers only around 125,000 pixels within the central $1^\circ$ of the visual field~\cite{ClarkVision2014,Smith2018}. Therefore, when viewing high-resolution real-world scenes, the human visual system adopts task-driven visual search strategies~\cite{wu2024v} to locate and analyze important regions. Following this biological principle, we aim to give large multimodal models (LMMs) a similar visual search ability by using visual grounding to focus on key image areas.

However, enabling LMMs to perform grounding-based visual reasoning remains challenging, mainly because acquiring accurate grounding annotations for standard visual question answering (VQA) datasets is both scarce and costly. These annotations are essential for constructing multi-turn grounding conversations used in supervised fine-tuning (SFT). In this paper, we demonstrate that \textit{accurate grounding behavior can emerge within a reinforcement learning framework, even when the training signal is limited to a binary reward derived solely from the correctness of the final answer}.

To this end, we introduce \textbf{Multi-turn Grounding-based Policy Optimization (MGPO)}, a reinforcement learning~(RL) framework built upon the Group Relative Policy Optimization (GRPO)~\cite{shao2024deepseekmath} algorithm with a novel multi-turn grounding-based rollout strategy. MGPO enables LMMs to iteratively focus on key image regions by automatically cropping sub-images based on model-predicted grounding coordinates within a multi-turn conversation. Given a high-resolution image and a question, the model first predicts the coordinates of key regions relevant to the query. An image cropping function is then triggered to extract and return the corresponding sub-image. In subsequent turns, the model can integrate previous in-context conversations (including both the original image and cropped sub-image) to solve the question. 


Empirically, we observe that LMMs struggle to autonomously trigger visual grounding during the rollout process. To mitigate the cold start problem without constructing additional annotated data, we design a fixed multi-turn conversation template: the first turn prompts the model to output relevant coordinates, while the second turn provides the sub-image and prompts the model to answer the question. To ensure stable optimization, 
policy loss is computed only on model outputs generated across multiple conversation rounds.

In summary, MGPO mainly offers the following advantages:

\begin{itemize}

\item \textbf{Top-down and Interpretable Visual Reasoning.} MGPO endows LMMs with a top-down, question-guided visual search mechanism tailored for high-resolution scenarios, enabling interpretable reasoning by indicating the image regions that should be attended to throughout the inference process.

\item \textbf{Overcomes Maximum Pixel Constraints.} MGPO overcomes the maximum pixel limitation of LMMs. Even when a high-resolution image must be resized to fit within pixel constraints, resulting in a blurred input, the model can still identify the relevant coordinates and extract clear sub-images from the original image for further analysis.

\item \textbf{Without Additional Grounding Annotations.} MGPO can be post-trained directly on standard VQA datasets without any extra grounding annotations. In contrast to closely related work such as DeepEyes~\cite{zheng2025deepeyes}, which still relies on ground-truth bounding boxes for data filtering and on intermediate grounding rewards during RL, MGPO is trained with the final-answer reward alone, yet still brings substantial improvements in intermediate grounding performance over GRPO~\cite{shao2024deepseekmath}.

\end{itemize}

\begin{figure*}[htb]
  \centering
  \includegraphics[width=1.0\textwidth]{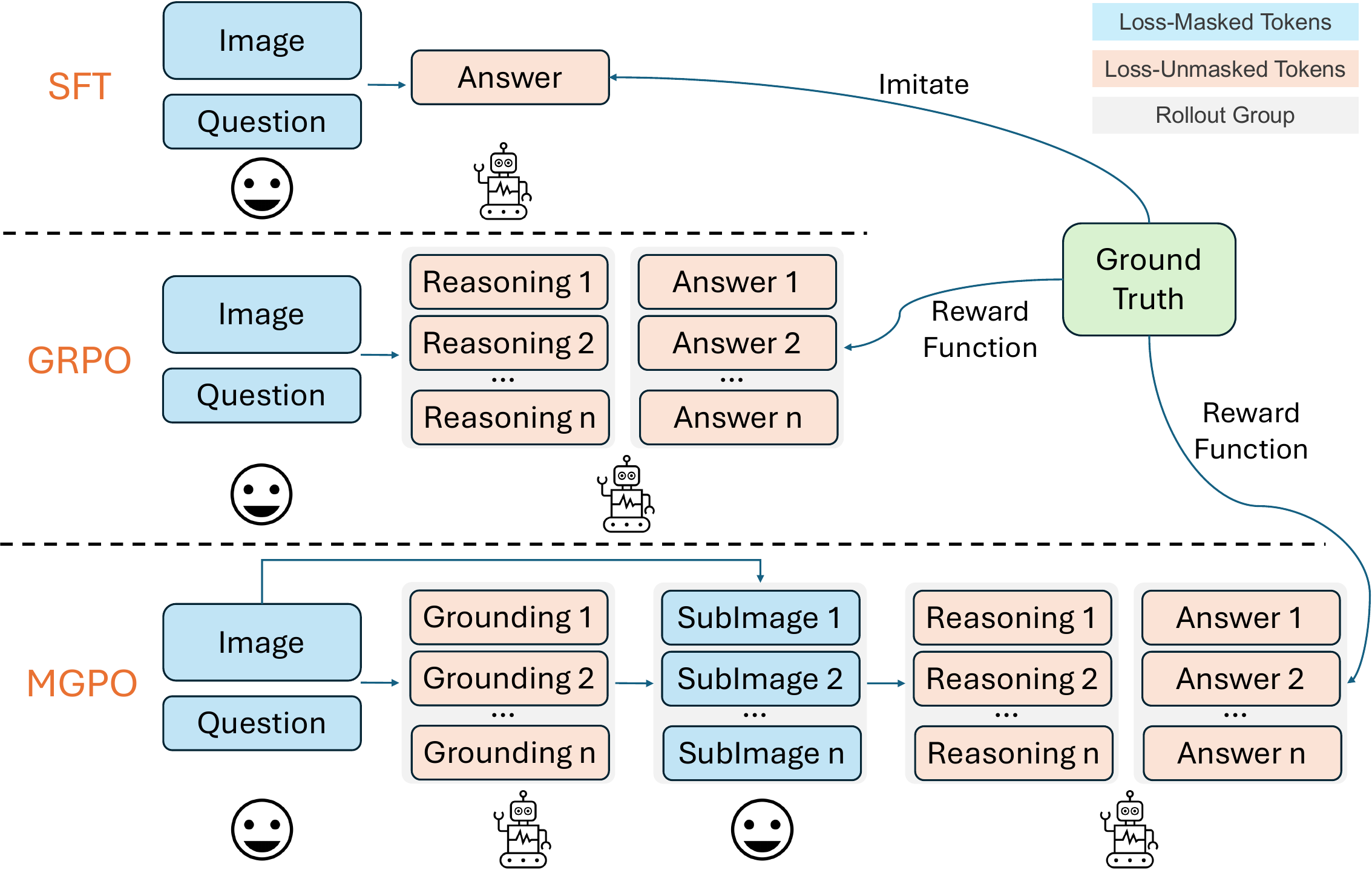}
	\centering
	\caption{Comparison of different post-training paradigms for LMMs. Our MGPO automatically crops and returns sub-image to the model based on its predicted grounding coordinates, enabling the model to iteratively focus on key regions and effectively solve high-resolution visual tasks.}
   \label{fig:comparison}
\end{figure*}

Ultimately, we use MGPO to post-train Qwen2.5-VL-7B~\cite{bai2025qwen25_vl} on visual-question-short answering data, and the resulting model achieves strong intermediate grounding performance without requiring grounding annotations. Compared to GRPO, MGPO yields a 5.4\% improvement on the in-distribution MME-Realworld~\cite{zhang2024mme} benchmark and a 5.2\% gain on the challenging out-of-distribution~(OOD) V* Bench~\cite{wu2024v}. Notably, leveraging with only 21K post-training samples, our model surpasses OpenAI’s o1 and GPT-4o models on the OOD V* Bench. 


\section{Related Work}

\subsection{Large Multimodal Models}
Recent advancements in LMMs~\cite{bai2025qwen25_vl,li2024llava-ov,liu2024oryx,liu2025ola,deitke2024molmo,hurst2024gpt,team2025kimi,chen2024expanding,chen2024far} have equipped the model with robust visual understanding capabilities, spanning from simple tasks such as common image comprehension to more challenging ones like long video analysis and mathematical reasoning. By leveraging high-quality instruction tuning data~\cite{li2024llava-ov,li2025eagle,deitke2024molmo}, enhanced model architecture~\cite{liu2024oryx,wang2024qwen2}, and meticulously designed training pipelines~\cite{liu2025ola,team2025kimi}, these models achieve fine-grained multi-modal alignment across various tasks. Despite these efforts, most LMMs concentrate on general visual understanding with common-resolution inputs, overlooking the inherent human ability to perceive fine-grained details in high-resolution images. Although some research~\cite{zhang2024beyond,guo2024llava,shi2025scaling} addresses high-resolution image analysis, it often falls short in generalization and competitive performance due to high training costs and insufficient data. In this work, we aim to tackle these challenges by developing LMMs capable of high-resolution image understanding with reduced training costs and minimal data requirements.

\subsection{Visual-Centric Multi-modal Reasoning} 
Multi-modal reasoning is a fundamental capability for large multimodal models (LMMs) when tackling complex and challenging tasks. Prior research has predominantly adopted the Chain-of-Thought (CoT) paradigm to either decompose intricate visual information into interpretable steps~\cite{liu2024chain,shao2024visual,sun2024visual} or enhance reasoning performance through instruction tuning~\cite{xu2024llava,dong2024insight,liu2024coarse}. More recently, inspired by the success of DeepSeek-R1~\cite{guo2025deepseek_r1}, a growing body of work~\cite{meng2025mm,huang2025vision,yang2025r1,zhang2025r1,peng2025skywork} has introduced reinforcement learning (RL) to further strengthen the reasoning abilities of LMMs. While these approaches substantially improve general reasoning across diverse tasks, they often underemphasize the ability to reason directly over visual content. Consequently, models may perform well on mathematical problem-solving, yet still exhibit weaknesses in perception-oriented tasks that require accurate visual grounding. A closely related line of work, exemplified by DeepEyes~\cite{zheng2025deepeyes}, explores ``thinking with images'' through RL, where the model iteratively grounds and crops sub-images to support reasoning. However, such methods still rely on ground-truth bounding boxes for data selection and on intermediate grounding rewards during RL. In this work, we address the same gap from a different angle: we develop a reinforcement learning framework that elicits accurate grounding behavior from the final-answer reward alone, removing the need for any grounding annotations.


\begin{figure*}[htb]
  \centering
  \includegraphics[width=0.98\textwidth]{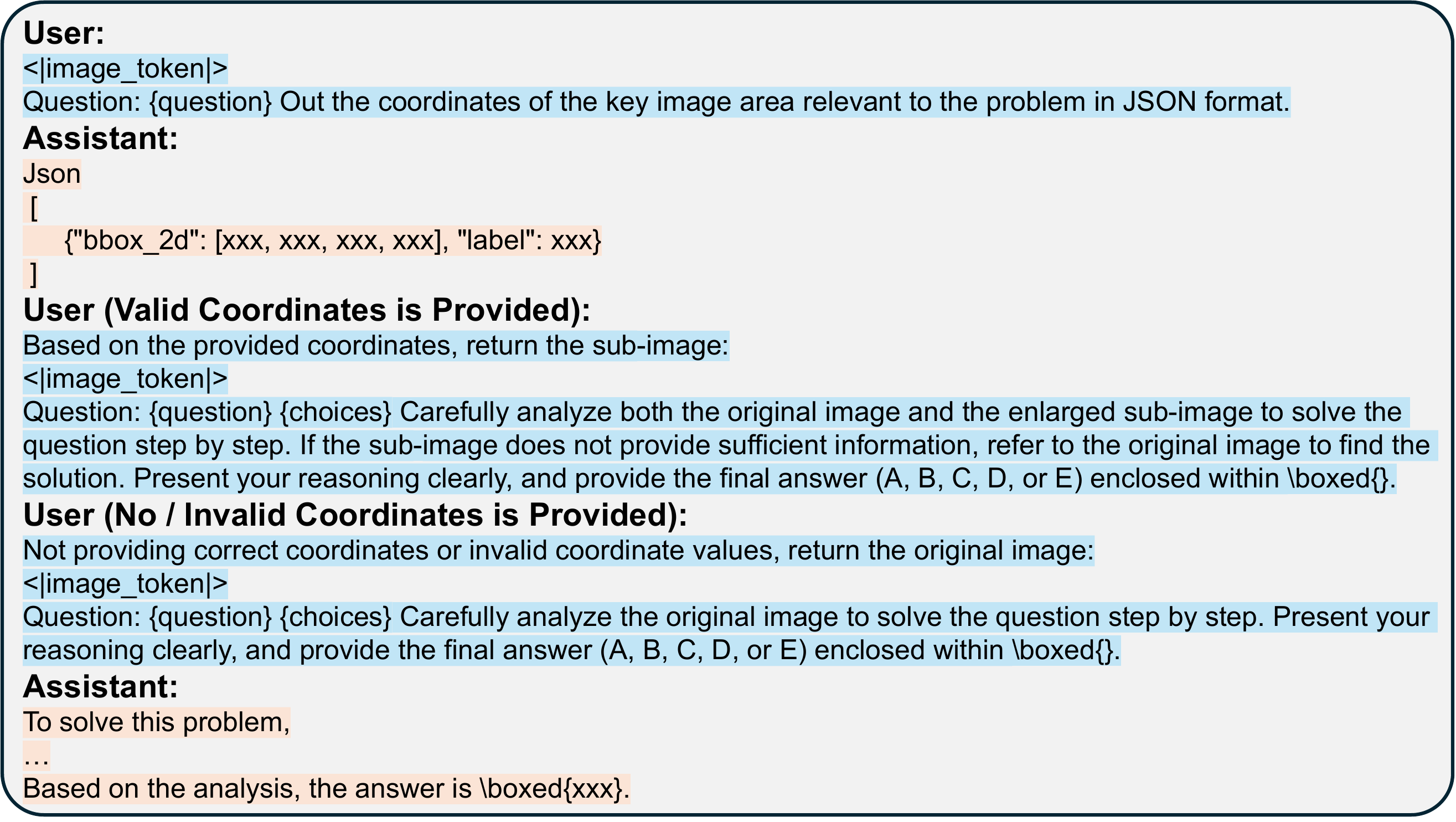}
	\centering
  \caption{Fixed multi-turn grounding template employed during MGPO training. This template guides the model to first identify the relevant regions and then perform reasoning to produce the final answer. When integrated with the MGPO training pipeline, this approach enables the model to develop grounded reasoning capabilities without a separate cold-start SFT stage. The bounding-box snippet in the upper-right corner is a raw output from the early stage of training, where the model has not yet learned to emit coordinates only; the format becomes increasingly clean as RL proceeds (see Figure~\ref{fig:format_reward}).}
   \label{fig:template}
   \vspace{-0.5em}
\end{figure*}

\section{Preliminaries}

\subsection{Advanced Baseline LMMs}

We adopt Qwen2.5-VL~\cite{bai2025qwen25_vl} as our baseline, which combines the Qwen2.5~\cite{yang2024qwen2.5} language model with a Native Resolution Vision Transformer (NaViT)~\cite{dehghani2023navit,tschannen2025siglip2} that processes images at their native resolutions. NaViT divides an input image $I$ into non-overlapping patches and aggregates groups of $m\times m$ patches via an MLP into visual tokens $X_i$, which are concatenated with textual tokens $X_t$ and jointly encoded for multimodal reasoning. Qwen2.5-VL also supports visual grounding by predicting bounding box coordinates $[x_1, y_1, x_2, y_2]$, though grounding typically requires explicit prompts and is not inherently triggered during reasoning.

\subsection{Single-Turn Multi-modal RL}

Figure~\ref{fig:comparison} compares post-training paradigms for LMMs. In supervised fine-tuning (SFT), the model directly imitates ground-truth responses, whereas reinforcement learning (RL) encourages exploration of reasoning trajectories that yield correct answers, going beyond the limits of imitation learning. We adopt Group Relative Policy Optimization (GRPO)~\cite{shao2024deepseekmath} as our RL baseline. Given visual tokens $X_i$ and textual context $X_t$, the model produces $G$ candidate outputs $\{X_a^g\}_{g=1}^{G}$. A rule-based reward function $\mathcal{R}$ assigns a binary score $r^g\in\{0,1\}$ indicating whether each prediction is correct. The policy gradient is:
\begin{equation}
\begin{split}
    \nabla_\theta \mathcal{J}_\mathrm{GRPO}(\theta) =  \mathbb{E}_{X_a^g \sim p_\theta} \\ \left[ (r^g - b) \cdot \nabla_\theta \log p_\theta(X_a^g \mid X_i, X_t) \right],
    \end{split}
\end{equation}
where $b = \frac{1}{G} \sum_{g=1}^G r^g$ is the average reward within the group. Only output tokens $X_a^g$ contribute to the loss, while input tokens serve solely as conditioning context.

\section{Multi-turn Grounding-Based RL}

\paragraph{Motivation.} While LMMs exhibit strong reasoning ability, they often fail to focus on the correct visual regions, which leads to inaccurate or hallucinated answers. When the model accurately localizes the regions referred to by the question, its reasoning becomes more coherent and the final answer more reliable; otherwise, misaligned attention frequently produces inconsistent reasoning steps. This observation motivates our multi-turn grounding-based RL strategy, which encourages the model to first attend to question-relevant regions before producing an answer, so as to better couple visual perception with reasoning in high-resolution scenarios.

\subsection{Formulation}

To address the challenges of high-resolution visual reasoning, we propose Multi-turn Grounding-based Policy Optimization (MGPO). \emph{MGPO is not a new RL optimizer; rather, it is an instantiation of GRPO with a multi-turn grounding-based rollout, where sub-image cropping is interleaved with the model's generation.} Figure~\ref{fig:comparison} shows the two-turn illustration compared with previous methods, and Algorithm~\ref{alg:mgpo} gives the general formulation. In this paradigm, the model operates over $K$ sequential interactions, dynamically grounding and reasoning by conditioning on the full history of visual and textual context at each step.

At each turn $k$, the model generates an output $X_a^{(k),g}$ based on the complete interaction history $\mathcal{H}^{(k)}$. If the model's output $X_a^{(k),g}$ contains grounding coordinates, an image cropping function is triggered to crop the relevant sub-image from the current visual input $X_i^{(k)}$, and this sub-image $X_i^{(k+1)}$ is provided as the new visual input for the next turn. Both the new sub-image and the model's output are appended to the history. The process continues until the model outputs a final answer, at which point the rollout terminates.

For each question, a group of $G$ rollouts is sampled. The reward for each rollout is computed based on the final answer, and a group baseline is used to reduce variance. The policy is optimized by maximizing the following objective:
\begin{equation}
\begin{split}
    \nabla_\theta \mathcal{J}_\mathrm{MGPO}(\theta) = 
    \mathbb{E}_{\{X_a^{(j),g}\} \sim p_\theta}
    \\ \left[
        (r^g - b) \cdot \sum_{j=1}^{k_g} \nabla_\theta \log p_\theta\left(X_a^{(j),g} \mid \mathcal{H}^{(j)}\right)
    \right],
\label{mgpo_queation}
\end{split}
\end{equation}
where $k_g$ denotes the number of steps in the $g$-th rollout, $r^g$ is the reward assigned to the final answer, $b$ is the average reward within the group, and $\mathcal{H}^{(j)}$ represents the complete interaction history up to step $j$. This formulation encourages the model to optimize all intermediate grounding and reasoning steps, which ultimately contribute to the correct final answer.

\begin{algorithm}[t]
\caption{Multi-turn Grounding-based Policy Optimization (MGPO)}
\label{alg:mgpo}
\KwIn{Policy model $\pi_\theta$; group size $G$}
\For{each rollout $g = 1, \ldots, G$}{
    Initialize $X_i^{(1)}$, $X_t^{(1)}$ from image and question\;
    $k \leftarrow 1$\;
    $\mathcal{H}^{(1)} = \{(X_i^{(1)}, X_t^{(1)})\}$\;
    \While{true}{
        Sample $X_a^{(k),g} \sim \pi_\theta(\cdot \mid \mathcal{H}^{(k)})$\;
        \If{$X_a^{(k),g}$ contains final answer}{
            \textbf{break}
        }
        \If{$X_a^{(k),g}$ contains grounding coordinates}{
            $X_i^{(k+1)} \leftarrow \mathrm{Crop}(X_i^{(k)},\, \mathrm{Coord}(X_a^{(k),g}))$\;
            $\mathcal{H}^{(k+1)} = \mathcal{H}^{(k)} \cup \{ (X_a^{(k),g},\, X_i^{(k+1)}) \}$\;
        }
        \Else{
            $\mathcal{H}^{(k+1)} = \mathcal{H}^{(k)} \cup \{ X_a^{(k),g} \}$\;
        }
        $k \leftarrow k+1$
    }
    Collect $\{X_a^{(j),g}\}_{j=1}^k$\;
}
Compute reward $r^g = \mathcal{R}(X_a^{(k),g})$ for each rollout $g$\;
Compute group average reward $b = \frac{1}{G} \sum_{g=1}^G r^g$\;
Update $\pi_\theta$ using policy gradients calculated by Eq.~\ref{mgpo_queation}
\end{algorithm}

\subsection{Implementation Details}

\vspace{5pt}
\noindent
\textbf{Multi-turn Template without Cold Start.}
In practice, we observe that LLMs struggle to autonomously generate grounding coordinates during the rollout process, which hinders effective multi-turn RL. To address this, we design a fixed two-turn dialogue template, as shown in Figure~\ref{fig:template}, to explicitly activate the model's grounding and reasoning abilities during the training process.

In the first turn, the model is prompted to output only the coordinates of the key image region relevant to the question. In the second turn, the image cropping function first checks the validity of the provided coordinates: if the coordinates are valid, the cropped sub-image is returned to the model; otherwise, the original image is returned. This template-based approach eliminates the cold-start stage, which would otherwise require constructing multi-turn grounding data and performing SFT before RL.

\noindent
\textbf{Grounding Key Visual Areas.} Within the two-turn MGPO framework, sub-images are extracted directly from the original high-resolution image. This design is particularly crucial when the original image resolution exceeds the maximum pixel limit of the LMM, as it enables the model to access higher-fidelity sub-images for processing.

Since the grounding coordinates predicted by Qwen2.5-VL are dependent on the resolution of the input image, it is necessary to normalize the predicted coordinates by the input image dimensions and subsequently map them back to the coordinate space of the original image:
\begin{equation}
    [\hat{x}_1, \hat{y}_1, \hat{x}_2, \hat{y}_2] = \frac{[x_1, y_1, x_2, y_2]}{S_\text{input}} \cdot S_\text{ori},
\end{equation}
where $S_\text{input}$ and $S_\text{ori}$ represent the width and height of the input and original images, respectively. A detailed illustration of this process is provided in the Appendix Figure~\ref{fig:framework}.




\begin{table*}[t]
        \vspace{-1.0em}
\caption{Performance comparison of different post-training paradigms for LMMs. V* Bench serves as an out-of-distribution~(OOD) evaluation, while MME-Realworld serves as an in-distribution~(ID) evaluation. Abbreviations: OCR—Optical Character Recognition in the wild; RS—Remote Sensing; DT—Diagram and Table; MO—Video Monitoring; AD—Autonomous Driving.}
\resizebox{\linewidth}{!}{
\begin{tabular}{llllllllllll}
\toprule
\multicolumn{1}{l|}{\multirow{3}{*}{Method}} & \multicolumn{1}{c|}{\multirow{3}{*}{\begin{tabular}[c]{@{}c@{}}V* Bench\\ (OOD)\end{tabular}}} & \multicolumn{10}{c}{MME-Realworld (ID)}                                                                                               \\ \cmidrule{3-12} 
\multicolumn{1}{l|}{}                        & \multicolumn{1}{c|}{}                                                                          & \multicolumn{1}{c|}{\multirow{2}{*}{Overall}} & \multicolumn{5}{c|}{Perception}                       & \multicolumn{4}{c}{Reasoning} \\
\cmidrule(lr){4-8}  \cmidrule(lr){9-12}
\multicolumn{1}{l|}{}                        & \multicolumn{1}{c|}{}                                                                          & \multicolumn{1}{c|}{}                         & OCR  & RS   & DT   & MO   & \multicolumn{1}{c|}{AD}   & OCR   & DT    & MO    & AD    \\ \midrule
\multicolumn{1}{l|}{Qwen2.5-VL-7B}           & \multicolumn{1}{l|}{58.6}                                                                      & \multicolumn{1}{l|}{46.1}                     & 75.6 & 33.3 & 65.0 & 30.1 & \multicolumn{1}{c|}{29.4} & 61.0  & 56.0  & 30.7  & 34.0  \\ \midrule
\multicolumn{12}{l}{\textit{+Post-Trianing (21K Samples):}}\\
\midrule
\multicolumn{1}{l|}{SFT}                     & \multicolumn{1}{l|}{71.7}                                                                          & \multicolumn{1}{l|}{58.7}                         & 85.6     &  \textbf{55.3}    &   78.0   &  43.6    &   \multicolumn{1}{c|}{43.7}     & 70.0      & 60.0      & 50.7     &   \textbf{41.0}    \\
\multicolumn{1}{l|}{GRPO}                    & \multicolumn{1}{l|}{71.2}                                                                      & \multicolumn{1}{l|}{55.1}                     & 81.6 & 51.3 & 75.0 & 42.9 & \multicolumn{1}{c|}{43.7} & 67.0  & 53.0  & 43.3  & 38.3  \\
\multicolumn{1}{l|}{MGPO}        & \multicolumn{1}{l|}{\textbf{76.4}\textcolor{mydarkgreen}{\bm{$_{+5.2}$}}}                                                                      & \multicolumn{1}{c|}{\textbf{60.5}\textcolor{mydarkgreen}{\bm{$_{+5.4}$}}}                     & \textbf{86.4} & 54.0 & \textbf{78.0} & \textbf{46.7} & \multicolumn{1}{c|}{\textbf{44.0}} & \textbf{74.0}  & \textbf{69.0}  & \textbf{52.7}  & 39.3  \\ \midrule
\multicolumn{12}{l}{\textit{OpenAI's Models:}}\\
\midrule
\multicolumn{1}{l|}{OpenAI o1}           & \multicolumn{1}{l|}{69.7}                                                                      & \multicolumn{1}{l|}{-}                     &- & - & - & - & \multicolumn{1}{c|}{-} & -  & -  & -  & -  \\ 
\multicolumn{1}{l|}{OpenAI GPT-4o}           & \multicolumn{1}{l|}{73.9}                                                                      & \multicolumn{1}{l|}{45.2}                     & 77.7 & 28.9 & 46.7 & 33.9 & \multicolumn{1}{c|}{22.4} & 61.4  & 44.8  & 36.5  & 26.4  \\ \bottomrule
\end{tabular}}
\label{tab:main_result}
\end{table*}

\section{Experiments}
\subsection{Datasets \& Metrics}

To evaluate the effectiveness of our approach, experiments are conducted on two established datasets: MME-Realworld~\cite{zhang2024mme} and V* Bench~\cite{wu2024v}. Both datasets are specifically designed to evaluate the capabilities of large multi-modal (LMMs) in analyzing high-resolution images and capturing fine-grained visual information.

\vspace{5pt}
\noindent
\textbf{MME-Realworld.~\cite{zhang2024mme}} The MME-Realworld dataset comprises a diverse array of tasks, which are systematically categorized into perception and reasoning domains. For in-distribution evaluation, the lite subset of MME-Realworld, consisting of 1,919 samples, is reserved as the test set, while the remaining 21,690 samples are utilized for training.


\vspace{5pt}
\noindent
\textbf{V* Bench.~\cite{wu2024v}} V* Bench serves as an out-of-distribution benchmark that focuses on detailed visual grounding on high-resolution images. This vision-centric benchmark requires LMMs to accurately localize and interpret specific visual information, which has also been adopted by OpenAI to assess the visual reasoning capabilities of their latest o3 and o4-mini models \cite{openai2024o3o4mini}. This benchmark contains 191 test samples.

\begin{figure*}[t]
    \centering
    \begin{minipage}[b]{0.32\textwidth}
        \centering
        \includegraphics[width=\textwidth,page=1]{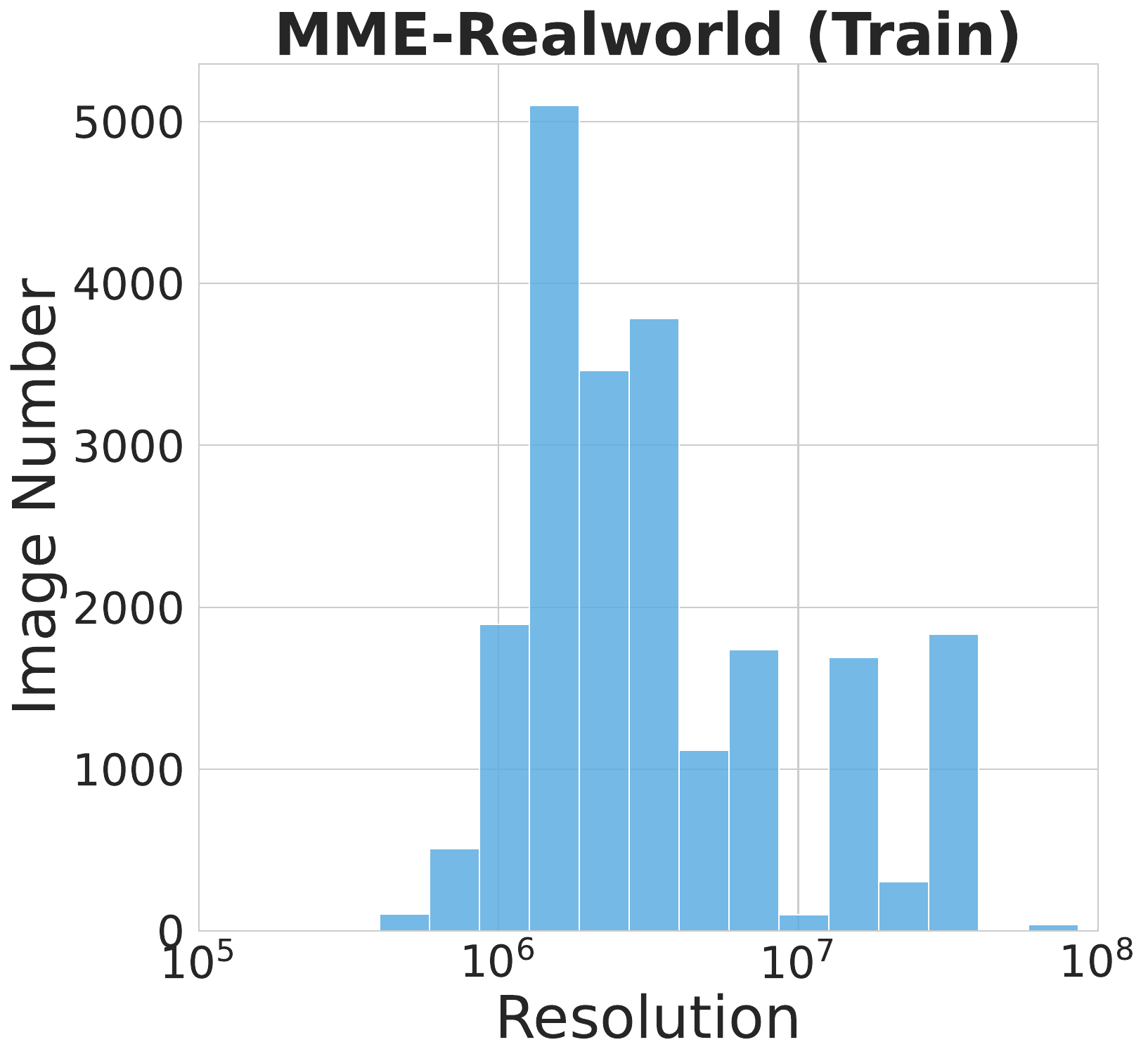}
    \end{minipage}
    \begin{minipage}[b]{0.32\textwidth}
        \centering
        \includegraphics[width=\textwidth,page=1]{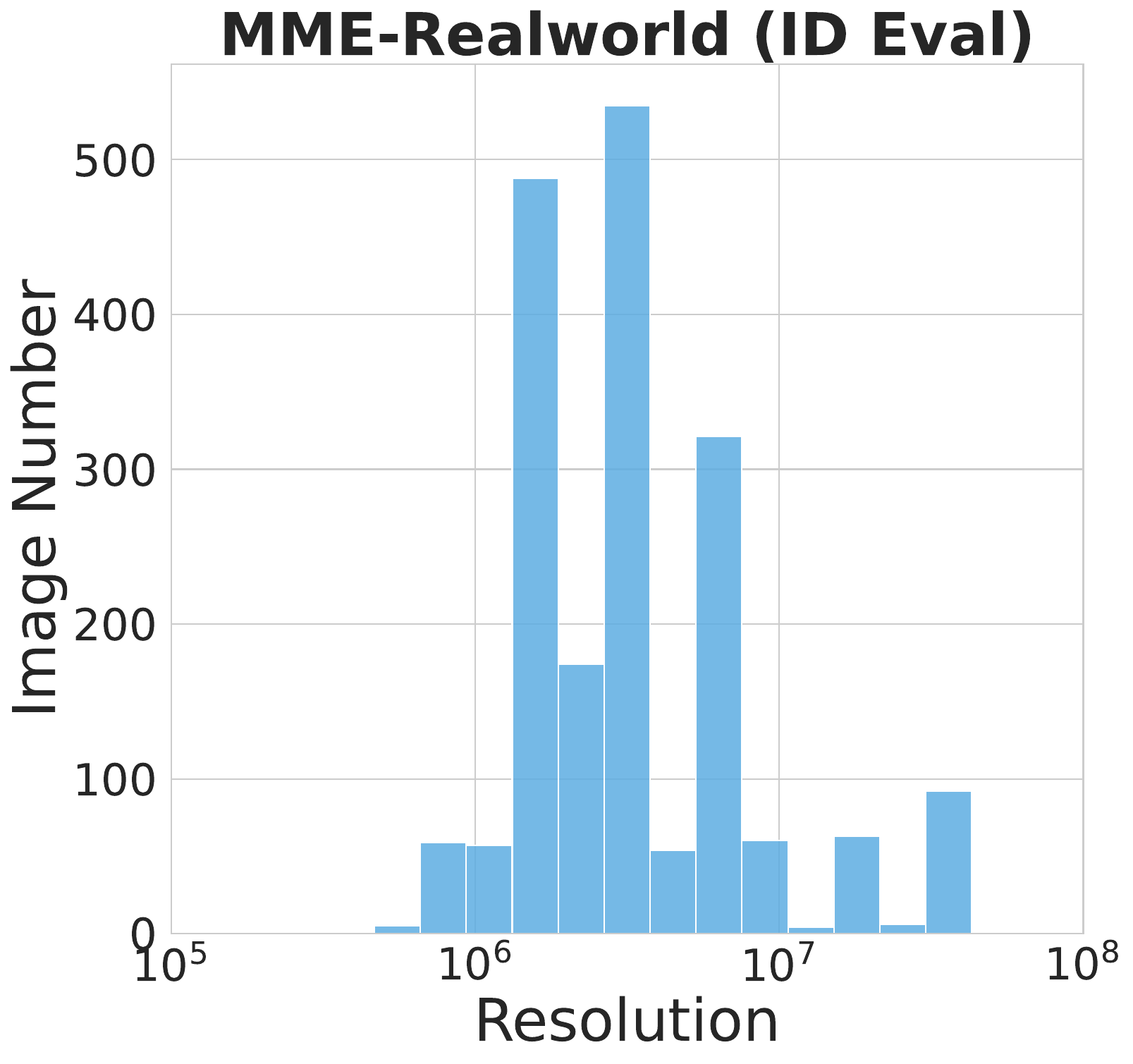}
    \end{minipage}
    \begin{minipage}[b]{0.32\textwidth}
        \centering
        \includegraphics[width=\textwidth,page=1]{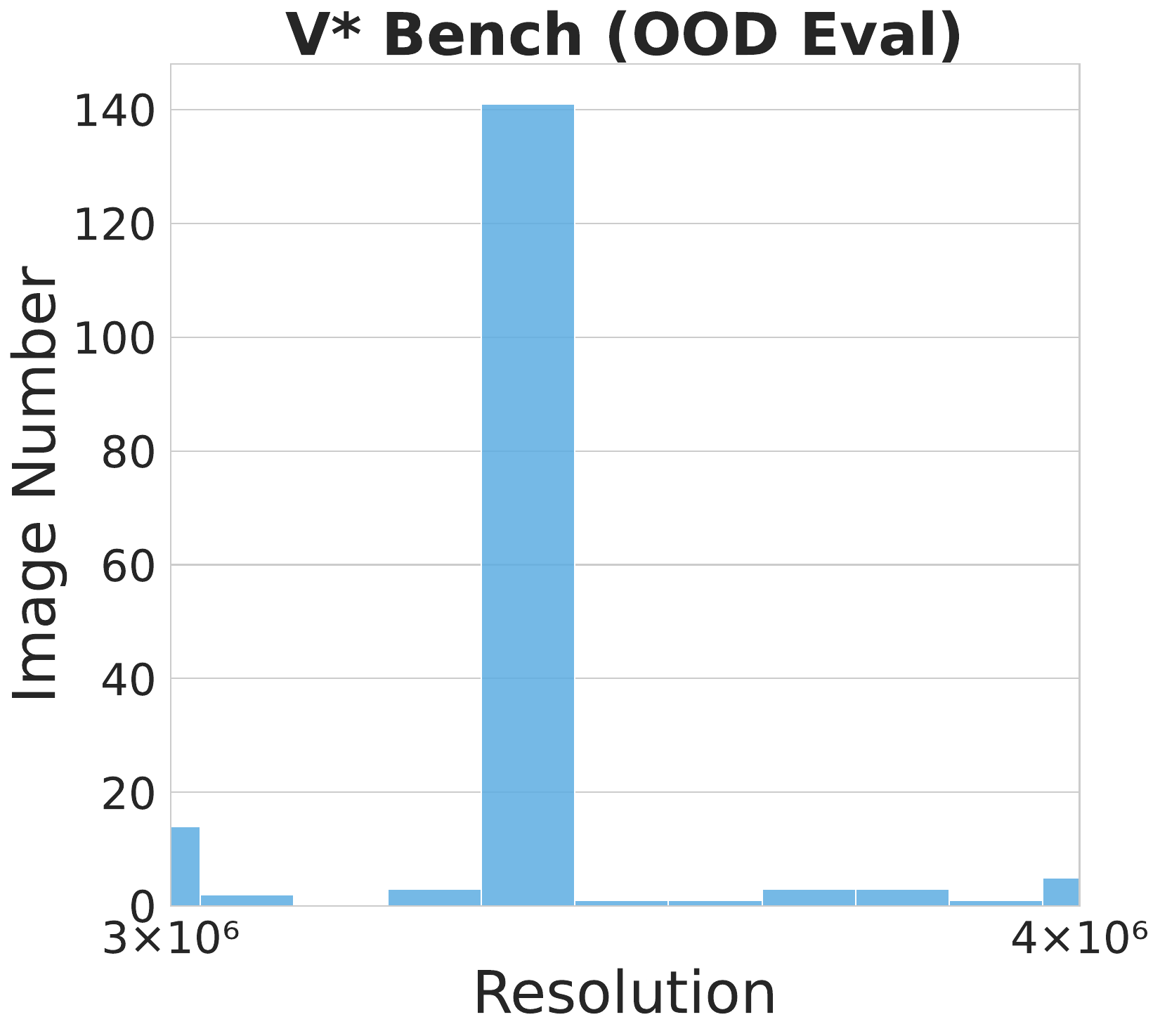}
    \end{minipage}
    \caption{Distribution of image resolutions (width $\times$ height) across different datasets.}
    \label{fig:resolution_distribution}
\end{figure*}

All datasets employ the multiple-choice question format, and model performance is consistently measured by accuracy on both the in-distribution (MME-Realworld) and out-of-distribution (V* Bench) test sets. Figure~\ref{fig:resolution_distribution} illustrates the distribution of image resolutions across different datasets.

\subsection{Experimental Setup}

We employ the verl~\cite{sheng2024hybridflow} framework to enable distributed training across multiple machines and GPUs, and utilize vLLM~\cite{kwon2023efficient} to accelerate inference during the rollout phase. For reinforcement learning, we adopt the naive GRPO~\cite{shao2024deepseekmath} algorithm as RL baseline, where a post-prompt is added: \textit{``\{question\}\textbackslash nOutput the coordinates of the key image area relevant to the problem in JSON format. And put the answer letter (A, B, C, D, or E) within \textbackslash boxed\{\}.'' } Both GRPO and our proposed MGPO leverage a binary accuracy reward function, assigning a reward of 1 if the final multiple-choice answer is correct and 0 otherwise.

All experiments are conducted using the Qwen2.5-VL-7B~\cite{bai2025qwen25_vl} model. To prevent out-of-memory errors, the maximum number of input image pixels is limited to 1,003,520 ($1280 \times 28 \times 28$), corresponding to a maximum of 1280 visual tokens per image. Images exceeding this pixel threshold are resized to comply with this constraint. More training details of MGPO are provided in Appendix~\ref{sec:imp-detail}.


\subsection{Main Results}

Table~\ref{tab:main_result} presents the performance comparison of different post-training paradigms on Qwen2.5-VL-7B, including SFT, GRPO and our MGPO. All three post-training methods substantially improve the model's performance on high-resolution visual tasks, as measured by both OOD V* Bench and ID MME-Realworld benchmarks.

Notably, we observe that GRPO does not yield significant improvements over SFT, which contrasts with conclusions drawn from prior work on multi-modal mathematical tasks~\cite{meng2025mm_eureka}. This abnormal result indicates that, for high-resolution vision-centric tasks, the primary challenge lies in enabling the model to perceive fine-grained image details, rather than performing complex, lengthy reasoning. 

In contrast, our MGPO framework demonstrates significant performance gains, consistently outperforming both SFT and GRPO across various benchmarks. Specifically, MGPO achieves a 5.2\% absolute improvement over the GRPO baseline on the V* Bench (OOD) benchmark and a 5.4\% increase in overall performance on the MME-Realworld (ID) dataset. These improvements underscore the effectiveness of the proposed multi-turn grounding and iterative sub-image cropping strategies in enhancing the model’s high-resolution visual understanding and reasoning abilities.

To further assess MGPO’s competitiveness, we compare our results with OpenAI’s o1~\cite{jaech2024openaio1} and GPT-4o~\cite{hurst2024gpt4o} models. For fairness, the comparison primarily focuses on the V* Bench (OOD) results. Remarkably, despite being built upon a relatively lightweight 7B model and trained on a modest dataset containing only 21k samples, our MGPO-post-trained model surpasses both o1 and GPT-4o in fine-grained visual perception. This highlights MGPO’s high data efficiency and strong generalization capacity under limited training resources.

\begin{figure}[t]
    \vspace{-1.0em}
    \flushleft
    \begin{minipage}[t]{0.49\textwidth}
        \flushleft
        \includegraphics[width=1.0\textwidth]{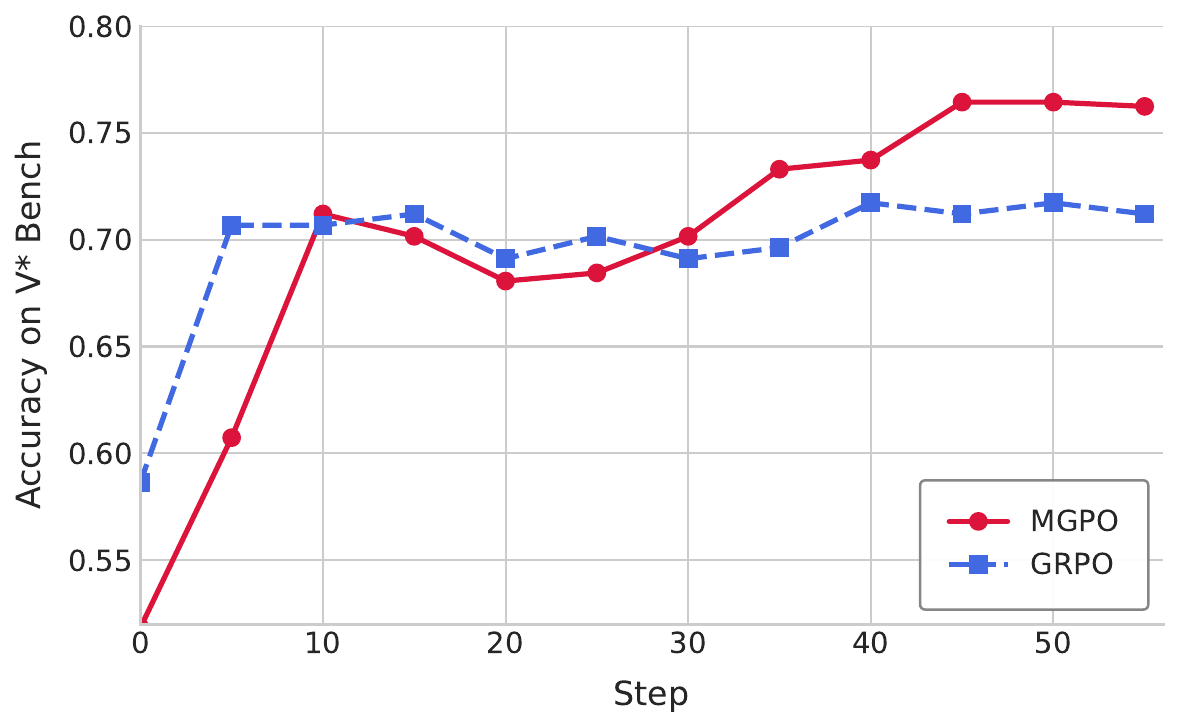}
        \caption{Performance comparison of V* Bench between our method (MGPO) and GRPO.}
        \label{fig:v_star}
    \end{minipage}
    \hfill
    \begin{minipage}[t]{0.49\textwidth}
        \flushright
        \includegraphics[width=1.0\textwidth]{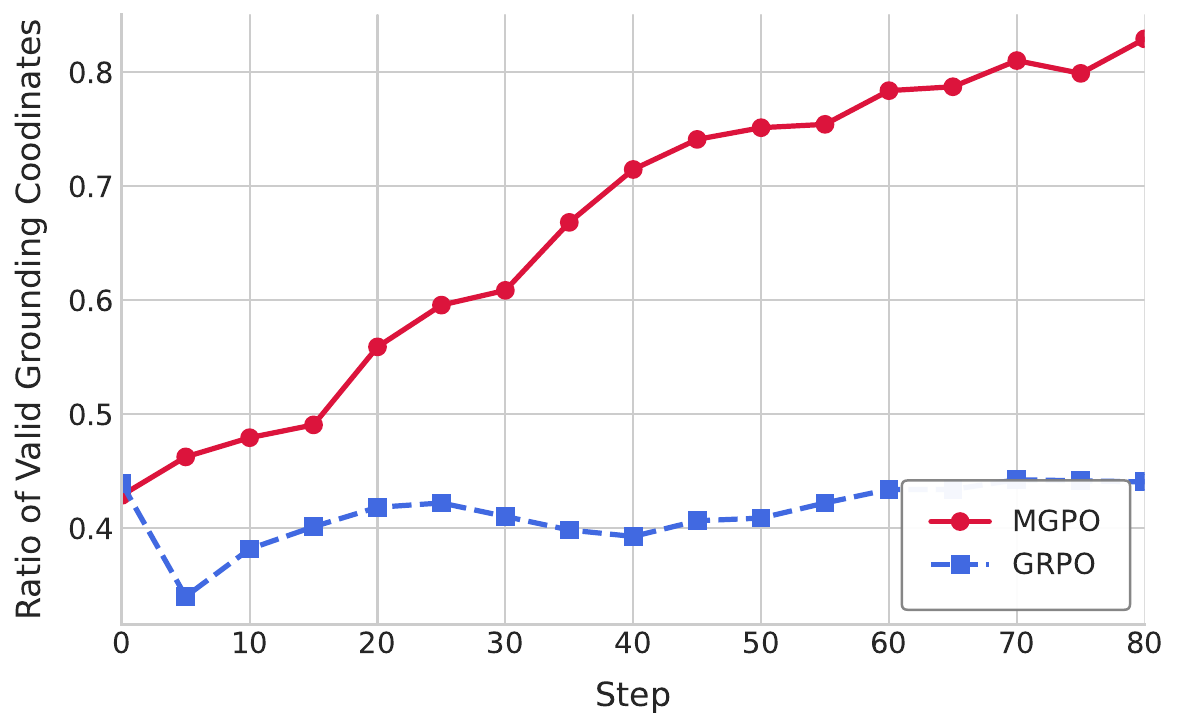}
        \caption{The ratio of valid grounding coordinates generated by models during RL rollouts. }
        \label{fig:format_reward}
    \end{minipage}
        \vspace{-1.0em}
\end{figure}

Figure~\ref{fig:v_star} further illustrates the comparative training trajectories of MGPO and GRPO on the V* Bench. Although MGPO exhibits lower accuracy at the initial stage, which is mainly due to the imperfect formulation of grounding information during early training, it quickly surpasses GRPO as learning progresses. This improvement highlights MGPO’s superior capability in addressing complex high-resolution visual scenarios that remain challenging for GRPO.

\vspace{5pt}
\noindent
\textbf{Generalization across Model Sizes and Diverse Benchmarks.}
To further examine how broadly MGPO applies, we extend the evaluation in two directions: (i) a smaller base model, Qwen2.5-VL-3B, to check whether the gains carry over to a different model scale; and (ii) two additional benchmark families, HR-Bench~\cite{wang2025hrbench} (4K and 8K) for high-resolution understanding and POPE~\cite{li2023pope} for object hallucination. The results are summarized in Table~\ref{tab:extra_benchmarks}. MGPO consistently outperforms the GRPO baseline on both 3B and 7B base models across all five benchmarks. The clear gains on HR-Bench suggest that multi-turn grounding effectively exploits fine-grained details in high-resolution images, while the improvements on POPE indicate that grounding answers in relevant visual evidence also helps reduce object hallucination.

\begin{table}[t]
\caption{Performance comparison on out-of-domain (V* Bench, HR-Bench), in-domain (MME-Realworld), and hallucination (POPE) benchmarks across two base model sizes.}
\centering
\footnotesize
\resizebox{\linewidth}{!}{
\begin{tabular}{llccccc}
\toprule
Base Model & Method & V* Bench & MME-RW & HR-4K & HR-8K & POPE \\
\midrule
\multirow{2}{*}{Qwen2.5-VL-3B} & GRPO & 66.5 & 50.3 & 67.9 & 64.9 & 82.5 \\
                               & MGPO & \textbf{67.0} & \textbf{52.2} & \textbf{70.9} & \textbf{68.1} & \textbf{83.2} \\
\midrule
\multirow{2}{*}{Qwen2.5-VL-7B} & GRPO & 71.2 & 55.1 & 69.8 & 66.3 & 84.5 \\
                               & MGPO & \textbf{76.4} & \textbf{60.5} & \textbf{74.2} & \textbf{71.2} & \textbf{86.1} \\
\bottomrule
\end{tabular}}
\label{tab:extra_benchmarks}
\end{table}

\vspace{5pt}
\noindent
\textbf{Comparison with DeepEyes.}
DeepEyes~\cite{zheng2025deepeyes} is a closely related work that also explores ``thinking with images'' via reinforcement learning. Table~\ref{tab:deepeyes} reports a head-to-head comparison on overlapping benchmarks. Two differences in the training setup are worth noting when reading the absolute numbers: (i) DeepEyes includes V*-style data in training, whereas for MGPO V* is held out as an out-of-distribution benchmark and never seen during training; and (ii) DeepEyes uses additional reward signals beyond final-answer correctness (e.g., format and grounding rewards that rely on bounding-box supervision), while MGPO is trained only with the final-answer accuracy reward and uses no grounding annotations. Even under this weaker supervision, MGPO remains competitive on HR-Bench 4K/8K, which is consistent with our main claim that strong grounding behavior can emerge from final-answer rewards alone.

\begin{table}[t]
\caption{Comparison with DeepEyes on overlapping benchmarks. V* Bench is OOD for MGPO (no V* data used in training), whereas DeepEyes integrates V*-style data and uses grounding rewards with bounding-box supervision.}
\centering
\footnotesize
\begin{tabular}{llccc}
\toprule
Base Model & Method & V* Bench & HR-4K & HR-8K \\
\midrule
\multirow{2}{*}{Qwen2.5-VL-7B} & MGPO     & 76.4 & 74.2 & 71.2 \\
                               & DeepEyes & 91.3 & 75.1 & 72.6 \\
\bottomrule
\end{tabular}
\label{tab:deepeyes}
\end{table}

\begin{table*}[htb]
\vspace{-1.0em}
\caption{Performance comparison of various post-training paradigms for LMMs under different maximum input image resolutions. GRPO struggles with high-resolution image inputs, performing even worse than SFT as the resolution grows. This highlights the inherent difficulty of LMMs in handling fine-grained visual information. In contrast, MGPO achieves notable improvements by effectively identifying and focusing on critical image regions.}
\centering
\footnotesize
\resizebox{\linewidth}{!}{
\begin{tabular}{lcccc}
\toprule
\multicolumn{1}{l|}{\multirow{3}{*}{Method}} & \multicolumn{4}{c}{Max Input Image Pixels}                                                     \\ \cmidrule{2-5} 
\multicolumn{1}{l|}{}                        & \multicolumn{2}{c|}{$640\times28\times28$}                           & \multicolumn{2}{c}{\textbf{$1280\times28\times28$}}      \\ \cmidrule{2-5} 
\multicolumn{1}{l|}{}                        & V* Bench (OOD) & \multicolumn{1}{c|}{MME-Realworld (ID)} & V* Bench (OOD) & MME-Realworld (ID) \\ \midrule
\multicolumn{1}{l|}{Qwen2.5-VL}              & 50.3           & \multicolumn{1}{c|}{40.9}               &    58.6            &   46.1                 \\ \midrule
\multicolumn{5}{l}{\textit{+Post-Training (21K Samples)}}                                                                                              \\ \midrule
\multicolumn{1}{l|}{SFT}                     & 66.5           & \multicolumn{1}{c|}{55.8}               & 71.7           & 58.7               \\
\multicolumn{1}{l|}{GRPO}                    & 64.4         & \multicolumn{1}{c|}{54.0}               & 71.2           & 55.1               \\
\multicolumn{1}{l|}{MGPO}                    &  \quad\;\,\; 73.3\color[HTML]{009901}\bm{$_{+8.9}$}          & \multicolumn{1}{c|}{\quad\;\,\; 57.0\color[HTML]{009901}\bm{$_{+3.0}$}}               & \quad\;\,\; 76.4\color[HTML]{009901}\bm{$_{+5.2}$}           & \quad\;\,\; 60.5\color[HTML]{009901}\bm{$_{+5.4}$}               \\ \bottomrule
\end{tabular}
}
\label{tab:different_max_pixels}
\vspace{-1.0em}
\end{table*}

\subsection{RL Brings Emergent Grounding Ability}

In this section, we highlight the insight that it is feasible to train powerful grounding-based RL models even without grounding annotations. This insight can broaden the applicability of grounding-based RL paradigms, as obtaining high-quality grounding annotations is often expensive and labor-intensive.

To assess whether models can develop accurate grounding capabilities in the absence of grounding supervision, we analyze the proportion of rollouts that generate valid grounding coordinates during RL training~(e.g., ensuring coordinates within the input image boundaries). Figure~\ref{fig:format_reward} illustrates the comparison between GRPO and MGPO. Regarding to GRPO, the ratio of valid grounding coordinates remains low and exhibits minimal improvement throughout training, indicating that the model struggles to ground correct image regions. In contrast, MGPO demonstrates a clear upward trajectory, with the proportion of valid grounding coordinates steadily increasing as training progresses. 

Additionally, we evaluate whether the grounded sub-images from the test set can be directly used to answer the question. As presented in Table~\ref{tab:valid_sub_imges}, a significantly higher proportion (65.2\%) of the sub-images generated by MGPO are sufficient for correctly answering the question, compared with 47.1\% for the original Qwen2.5-VL and 50.2\% for the GRPO baseline. This indicates that MGPO effectively identifies highly informative regions, even though no auxiliary reward is provided for generating valid sub-image coordinates: the model autonomously learns to localize key regions and utilize sub-images to improve question answering.

\begin{table}[t]
    \caption{Ratio of grounded sub-images that can directly answer the question on V* Bench.}
    \centering
    \footnotesize
    \begin{tabular}{lc}
    \toprule
    Method     & Ratio (\%) \\
    \midrule
    Qwen2.5-VL &  47.1 \\
    \;+GRPO    &  50.2 \\
    \;+MGPO    &  \textbf{65.2} \\
    \bottomrule
    \end{tabular}
    \label{tab:valid_sub_imges}
\end{table}

\vspace{5pt}
\noindent
\textbf{Robustness to Imperfect Grounding.}
Although MGPO consistently improves grounding quality, its final-answer correctness does not critically depend on perfect grounding at every step. The two-turn template provides a built-in safety net: in the answering turn the model still has access to the original global image from the first turn, so when a cropped sub-image is uninformative or inaccurate, it can fall back on the global context to produce an answer. In the worst case, MGPO is therefore no worse than the standard Qwen2.5-VL baseline that uses only the full image, while it benefits from focused sub-images whenever grounding is reliable.

We further conduct experiments on the image counting task in Appendix~\ref{sec:img-count}, leveraging the fact that the image count dataset provides both the grounding annotations and the corresponding count as the final answer. We compare two reward functions for RL post-training: (1) the binary accuracy reward based solely on the correctness of the final count, and (2) incorporating an additional point reward based on grounding annotations. Both qualitative and quantitative results indicate that introducing the additional point reward does not yield significant performance improvements, supporting our design choice of relying solely on the final-answer reward.







\vspace{5pt}
\noindent
\textbf{Effect of Maximum Input Image Resolution.} 
Table~\ref{tab:different_max_pixels} compares the impact of varying maximum input image resolutions for LMMs. We observe that MGPO yields greater performance improvements on the V* Bench when the maximum input pixel limit is lower. This is because, when high-resolution images are aggressively resized, many tasks become more challenging to solve directly. However, MGPO can first identify key regions and crop clearer sub-images from the original image, which enables more effective task completion under constrained resolution settings.

\vspace{5pt}
\noindent
\textbf{Zero-Shot Generalization to More Interaction Turns.}
Although MGPO is trained with a fixed two-turn template for stability, the trained model is not tied to that interaction length at inference. We evaluate the trained 7B model on V* Bench by allowing one additional grounding round at test time, without any further training. As shown in Table~\ref{tab:turn_generalization}, extending the interaction from 2 to 3 turns yields a further improvement (76.4 $\rightarrow$ 77.2), suggesting that MGPO has learned a transferable grounding-then-reasoning behavior rather than overfitting to the two-turn template.

\begin{table}[h]
\caption{Zero-shot generalization to a 3-turn interaction at inference time on V* Bench (Qwen2.5-VL-7B + MGPO).}
\centering
\footnotesize
\begin{tabular}{lc}
\toprule
Evaluation Setting & V* Bench \\
\midrule
2-turn (training setup)            & 76.4 \\
3-turn (zero-shot at inference)    & \textbf{77.2} \\
\bottomrule
\end{tabular}
\label{tab:turn_generalization}
\end{table}




\section{Conclusion}

This paper introduces MGPO, a multi-turn grounding-based reinforcement learning framework built on GRPO with a grounding-aware rollout strategy, which elicits a thinking-with-images behavior for high-resolution real-world scenarios. Using only a binary reward derived from the correctness of the final answer, MGPO enables the model to develop robust grounding abilities during RL training, without requiring any grounding annotations. We hope the insights from MGPO can further advance research on grounding-based visual reasoning and contribute to more generalizable vision--language models.

\section*{Limitations}
All experiments of MGPO are conducted using a fixed two-turn template during training, rather than allowing the model to autonomously decide when to perform image cropping based on the input question, as illustrated in lasted OpenAI models such as o3 and o4-mini~\cite{openai2024o3o4mini}. This limitation stems from our observation that Qwen2.5-VL~\cite{bai2025qwen25_vl}, when directly subjected to RL post-training, struggles to generate grounding coordinates without explicit prompt guidance. That said, the trained model is not strictly tied to the two-turn setting at inference: as reported in Table~\ref{tab:turn_generalization}, it can zero-shot generalize to a 3-turn interaction and obtain a further improvement on V* Bench, suggesting that the learned grounding-then-reasoning behavior is not merely tied to the fixed template.

\noindent Nevertheless, we believe that our trained models can be leveraged to generate high-quality chain-of-thought (CoT) data for subsequent SFT. By adopting a multi-stage training strategy that combines SFT and RL, as in DeepSeek-R1~\cite{guo2025deepseek_r1}, may ultimately enable the model to autonomously decide when and how to perform grounding. We leave this direction for future work.

\section*{Ethical Considerations}

This research adheres to the ethical standards and guidelines for responsible AI research. The proposed MGPO framework and all associated experiments are conducted using publicly available datasets that do not contain personally identifiable information or sensitive content. All dataset licenses are respected, and appropriate citation and use policies are followed.

\noindent The development of MGPO aims to advance grounding-based visual reasoning rather than enable harmful or deceptive uses of AI technologies. The authors recognize the potential social and ethical implications of LMMs, including possible biases inherited from training data and unintended misuse in sensitive domains. To mitigate these risks, evaluation and analysis procedures are transparently reported, and results are carefully interpreted within the intended research scope. The authors encourage future work to continue improving explanability and transparency in grounding-based reinforcement learning systems.


\bibliography{acl_latex}

\clearpage
\appendix

\section*{Supplementary}
In the supplementary material, we first present additional implementation and training details of the proposed MGPO framework in Section~\ref{sec:imp-detail}. After that, we include an LLM usage section in Section~\ref{sec:llm_use}, detailing how we use LLMs properly in this paper. Then, in Section~\ref{sec:img-count}, we provide further insights into the optimization and behavioral characteristics of grounding-based reinforcement learning models through experiments on image counting tasks. Finally, Section~\ref{sec:quality} reports extended qualitative results of the MGPO framework, offering a more comprehensive understanding of its performance across various visual reasoning scenarios.

\section{Training Details}
\label{sec:imp-detail}
Model training is conducted on a computational cluster comprising four nodes, each equipped with eight H100 GPUs. Optimization is performed using the AdamW~\cite{loshchilov2017decoupled} optimizer with a fixed learning rate of $1 \times 10^{-6}$. The global training batch size is set to 512, and a mini-batch size of 128 is used for each iteration. During rollout, eight responses are sampled per prompt. The Proximal Policy Optimization (PPO) clip ratio~\cite{schulman2017proximal}, which constrains the magnitude of policy updates and plays a role analogous to the learning rate in reinforcement learning, is set to the default value of 0.2. We strictly adhere to the Apache-2.0 License of the training data from MME-RealWorld~\cite{zhang2024mme}. We use LMMs-Eval~\cite{lmms_eval2024} to evaluate our model.

\section{LLM Usage}
\label{sec:llm_use}
In this paper, we clarify that Large Language Models (LLMs) are employed solely to support and refine the writing process. Specifically, we use LLMs to provide sentence-level suggestions and to enhance the overall fluency of the text.

\section{Further Experiments on Image Counting Tasks}
\label{sec:img-count}
To further substantiate the insight that training powerful grounding-based RL models even without grounding annotations, we conduct additional experiments on the image counting task. Specifically, we randomly sample 3,000 instances from the Pixmo-Points~\cite{deitke2024molmo} dataset for post-training, which provides both the grounding annotations (in point format) and the corresponding count as the final answer. Pixmo-Count~\cite{deitke2024molmo} is used as the in-distribution (ID) evaluation benchmark, while FSC-147~\cite{ranjan2021fsc_147} serves as the out-of-distribution (OOD) benchmark.


\begin{figure}[htb]
    \centering
    \includegraphics[width=\linewidth]{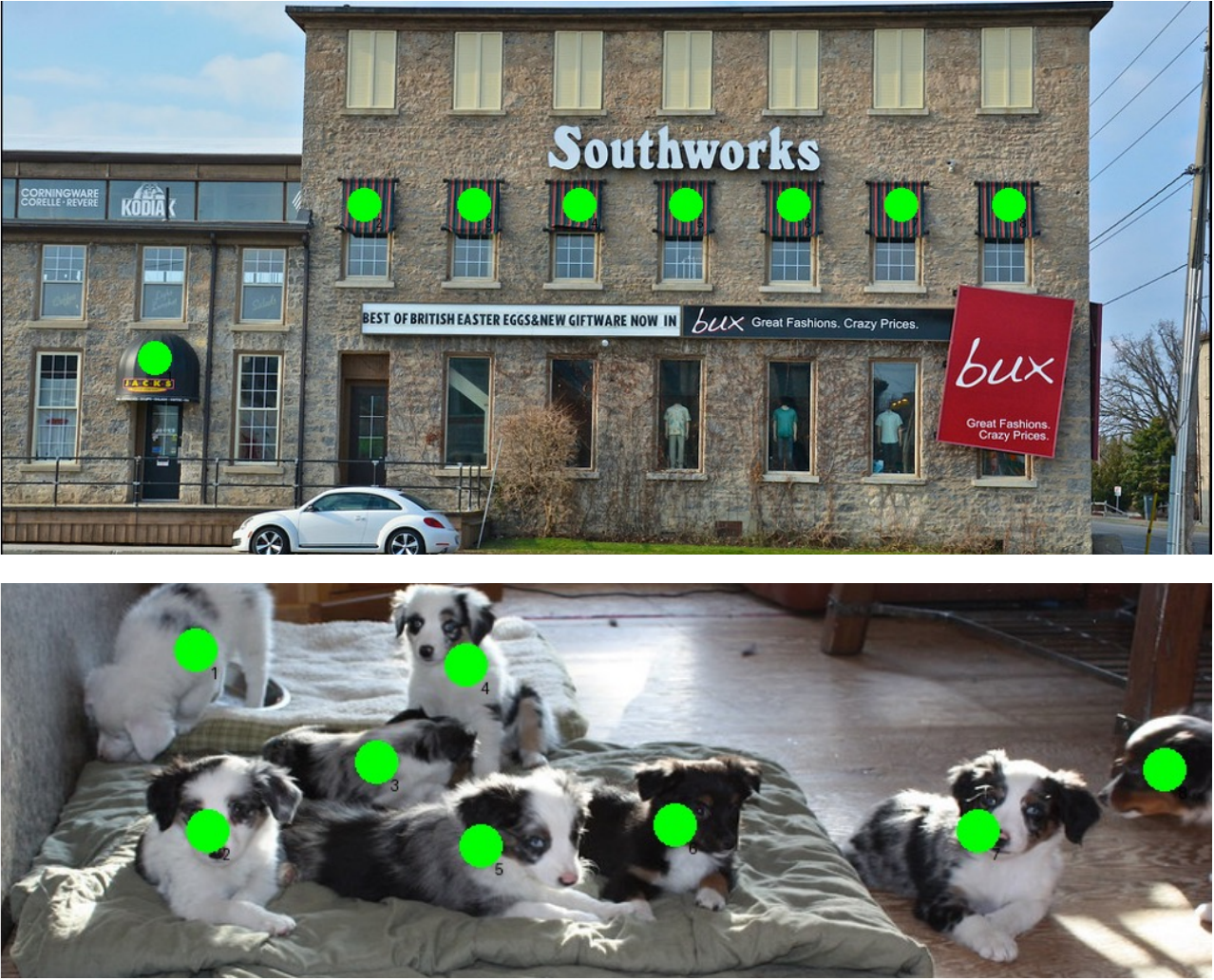}
    \caption{Visualization of point predictions from the GRPO model trained with only an accuracy reward.}
            \label{fig:count_examples}
\end{figure}

\begin{table}[htb]
  \centering
  \caption{Performance comparison of the image count task. Additional point rewards do not lead to significant performance improvements.}
  \resizebox{\linewidth}{!}{
\begin{tabular}{c|cc}
\midrule
\multirow{2}{*}{Method} & \multirow{2}{*}{\begin{tabular}[c]{@{}c@{}}FSC-147\\ (OOD)\end{tabular}} & \multirow{2}{*}{\begin{tabular}[c]{@{}c@{}}Pixmo-Count\\ (ID)\end{tabular}} \\
                        &                                                                          &                                                                             \\ \hline
\multicolumn{1}{l|}{Qwen2.5-VL-7B}                  & 59.9        & 13.0    \\ \hline
\multicolumn{3}{l}{+Post-Trianing (3K Samples)}                             \\ \hline
\multicolumn{1}{l|}{SFT}                            & 72.7        & 24.1    \\
\multicolumn{1}{l|}{GRPO {\footnotesize(Acc Reward)}}         & 81.0        & 35.0    \\
\multicolumn{1}{l|}{GRPO {\footnotesize(Acc + Point Reward)}} & 81.9        & 34.9    \\ \midrule
\end{tabular}
\label{tab:count_results}
}
\end{table}

During GRPO post-training, the model is prompted to first grounding (point) each object in the image and subsequently provide the total count. We compare two reward function: (1) the binary accuracy reward based solely on the correctness of the final count, and (2) incorporating an additional point reward. The point reward is computed by matching the model's predicted point list with the ground-truth point list using the Hungarian algorithm~\cite{kuhn1955hungarian}, such that a higher ratio of match ratio results in a higher reward.

\noindent The results, summarized in Table~\ref{tab:count_results}, indicate that introducing the additional point reward does not yield significant performance improvements. We further visualize the outputs of the GRPO model trained solely with accuracy reward (see Figure~\ref{fig:count_examples}), and observe that the model is capable of accurately localizing object points even without explicit grounding supervision. These results support our conclusion that explicit grounding annotations are not necessary for effective RL-based learning, as the model inherently learns to perform precise grounding as a prerequisite for solving counting tasks.

\section{Qualitative Results}
\label{sec:quality}
\begin{figure*}[t]
  \centering
  \includegraphics[width=1.0\textwidth]{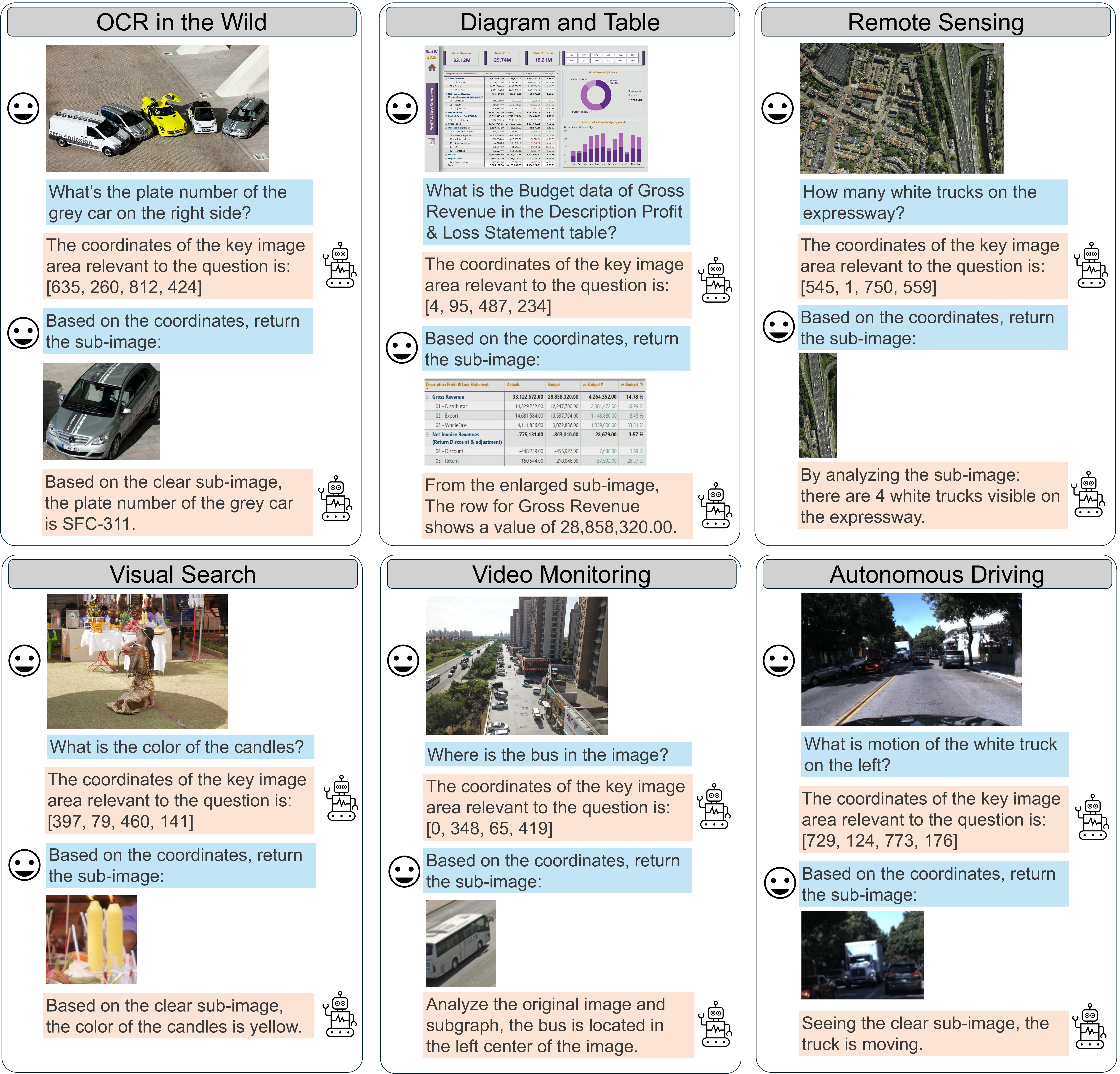}
	\centering
	\caption{Examples of models trained with multi-turn grounding-based RL on high-resolution real-world tasks. The model first identifies key regions, which are then automatically cropped and returned as sub-images. Notably, despite only a binary reward function derived from the correctness of the final answer, the model gradually emerges with robust grounding capability throughout the RL process. The conversation in the figure only shows key parts, the full conversation is provided in Appendix Figure~\ref{fig:specific_outputs}.}
   \label{fig:example-result}
\end{figure*}

\begin{figure*}[t]
  \centering
  \includegraphics[width=1.0\textwidth]{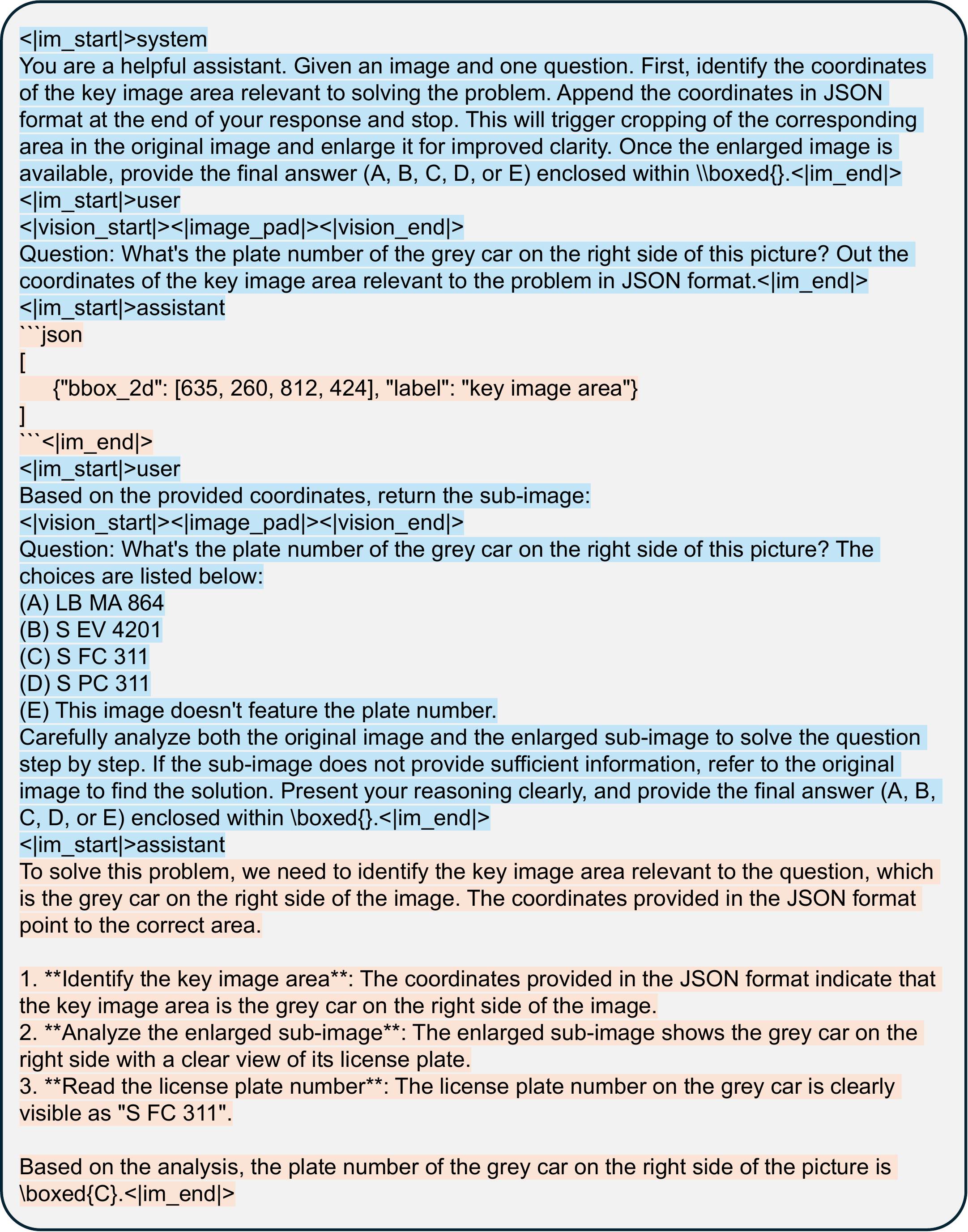}
	\centering
	\caption{A full conversation example of the MGPO post-trained model on high-resolution image tasks.}
   \label{fig:specific_outputs}
\end{figure*}

We provide more qualitative results to demonstrate the effectiveness of MGPO in Figure~\ref{fig:example-result} and Figure~\ref{fig:specific_outputs}. We visualize both the grounding results and final answer. MGPO enables the model to emerge with accurate grounding capabilities with only a binary reward, leading to the final correct answer.

\begin{figure*}[t]
  \centering
  \includegraphics[width=1.0\textwidth]{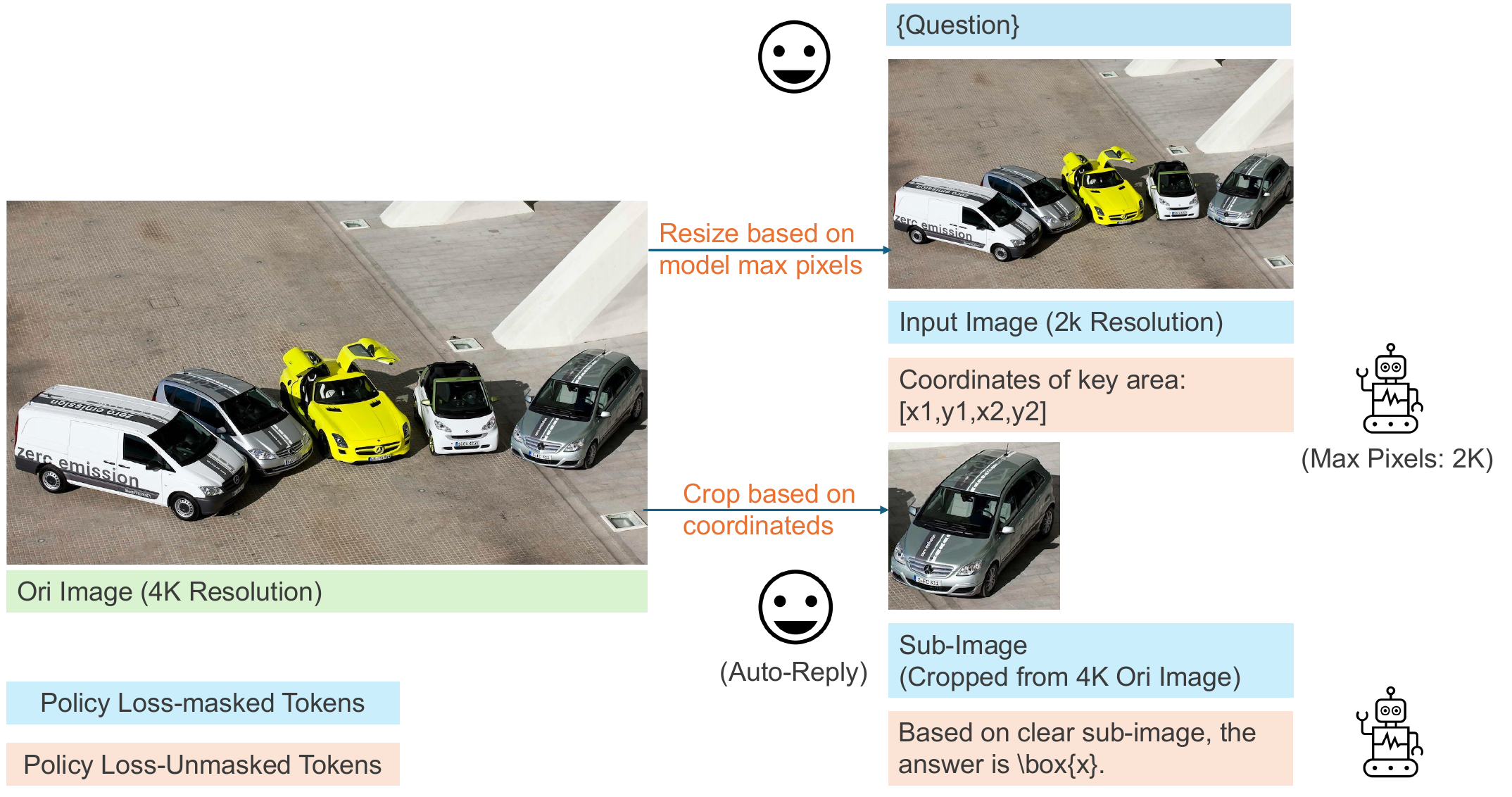}
	\centering
	\caption{An illustration of cropping a sub-image based on grounding coordinates. }
   \label{fig:framework}
\end{figure*}

\end{document}